\documentclass[journal,transmag]{IEEEtran}

\usepackage{cite}
\usepackage{amsmath,amssymb,amsfonts}
\usepackage{graphicx}
\usepackage{textcomp}
\usepackage{xcolor}
\usepackage{tikz}
\usetikzlibrary{positioning,arrows.meta}
\usepackage{subcaption}
\usepackage{booktabs}
\usepackage[colorlinks,urlcolor=blue,linkcolor=blue,citecolor=blue]{hyperref}
\usepackage{tabularx}
\usepackage{booktabs}
\usepackage{subcaption}
\usepackage{xurl} 
\usepackage{subcaption}
\begin{document}

\title{Bridging LLM Agents and Data Spaces: An Architectural Mediation Approach using the Model Context Protocol}

\author{\IEEEauthorblockN{Jaime Alonso Ruiz\IEEEauthorrefmark{1},
Carlos Aparicio\IEEEauthorrefmark{1},
Gabriel Huecas\IEEEauthorrefmark{1}, 
Joaquín Salvachúa\IEEEauthorrefmark{1}, and
Andres Munoz-Arcentales\IEEEauthorrefmark{1}}
\IEEEauthorblockA{\IEEEauthorrefmark{1}Information Processing and Telecommunications Center,\\
Escuela Técnica Superior de Ingenieros de Telecomunicación,\\
Universidad Politécnica de Madrid, Madrid, Spain}
\thanks{Corresponding author: Andres Munoz-Arcentales (email: joseandres.munoz@upm.es).}}

\markboth{}%
{Alonso Ruiz \MakeLowercase{\textit{et al.}}: Bridging LLM Agents and Data Spaces}

\IEEEtitleabstractindextext{%
\begin{abstract}
Data Spaces enable sovereign and governed data sharing across organizational boundaries, but their integration with AI agents remains challenging due to mismatches between probabilistic language model interactions and policy-driven data infrastructures. This article presents an architectural mediation approach based on the Model Context Protocol (MCP), implemented through the Eunomia Agent, to enable controlled interaction between large language model (LLM) agents and data space services. The proposed mediation layer translates data space capabilities into structured, schema-driven tools that AI agents can discover and invoke while preserving governance constraints. A prototype implementation validates end-to-end interaction across catalog discovery, metadata retrieval, and data service invocation without modifying existing data space  components. Results demonstrate that protocol-based mediation enables interoperable and standards-aligned integration of AI agents into data space  ecosystems. The approach provides practical guidance for organizations seeking to introduce AI-driven automation into governed data-sharing environments while maintaining compliance, interoperability, and architectural separation of concerns.
\end{abstract}

\begin{IEEEkeywords}
Data Spaces, Large Language Models, Model Context Protocol, Architectural Mediation, Data Governance, Interoperability.
\end{IEEEkeywords}}

\maketitle
\IEEEdisplaynontitleabstractindextext
\IEEEpeerreviewmaketitle

\section{Introduction}

\IEEEPARstart{L}{arge} Language Models (LLMs) are increasingly deployed as autonomous agents capable of discovering, selecting, and invoking external tools and data services \cite{ref1,ref2}. However, integrating LLM-based agents with structured data infrastructures remains challenging, particularly in domains governed by stringent interoperability, trust, and policy requirements.

Data Spaces have emerged as a prominent paradigm for secure, decentralized, and interoperable data sharing across organizational boundaries \cite{ref17}. Positioned by the European Commission as a cornerstone of its digital transformation strategy \cite{ref3,ref4}, Data Spaces aim to enable sovereign data sharing through standardized architectures such as the IDS. The International Data Spaces Association (IDSA) has developed comprehensive technical standards, including the IDS Reference Architecture Model and the data space  Protocol, currently undergoing ISO/IEC standardization \cite{ref5,ref6}. These frameworks define mechanisms for catalog publication, identity verification, contract negotiation, policy enforcement, and controlled data transfer. The Eclipse data space  Components (EDC) framework provides a reference implementation deployed in real-world Data Spaces such as Catena-X \cite{ref7}. However, implementations such as the Eunomia Agent also operationalize these principles by providing catalog services, metadata management, negotiation workflows, and integration with data services in accordance with established standards.

Despite the complementary objectives of AI-driven orchestration and sovereign data exchange, LLM agents and data space  infrastructures exhibit structural misalignment. LLMs operate through declarative, schema-defined interaction models in which tools are discovered dynamically and invoked with structured parameters. Data Spaces, in contrast, rely on procedural, protocol-bound workflows designed for explicit machine-to-machine integration. Directly exposing data space  APIs to LLM agents risks brittle coupling, misuse of protocol semantics, and potential violations of governance constraints.

The Model Context Protocol (MCP), recently proposed by Anthropic as a standardized mechanism for exposing tools to LLMs in a model-agnostic manner \cite{ref8}, offers a promising solution. MCP defines a structured pattern where external systems expose capabilities as typed tools, enabling LLMs to discover and invoke functionalities dynamically. With support from major providers including OpenAI and Google DeepMind, MCP appears suitable for bridging LLM agents and complex backend systems.

Despite this potential, limited work explores how MCP can be applied to real data space  implementations without compromising core principles of data sovereignty and policy enforcement. This gap is critical given emerging European data regulations such as the Data Act and Data Governance Act \cite{ref9,ref10}.

This work addresses this integration gap by proposing an Architectural Mediation Layer based on the Model Context Protocol. Rather than modifying the Eunomia Agent or embedding AI-specific logic into the data space  platform, the mediation layer operates externally. It exposes selected data space  capabilities as MCP tools and translates LLM-driven invocations into compliant API interactions. In doing so, it preserves data space  integrity while enabling structured interoperability with LLM agents.

The contributions of this paper are fourfold:
\begin{enumerate}
    \item We formalize an architectural approach for mediating LLM interaction with protocol-driven data space  infrastructures.
    \item We implement an MCP-compliant mediation server exposing catalog discovery and data service invocation capabilities of the Eunomia Agent.
    \item We validate feasibility through deterministic evaluation and integration with a production MCP client.
    \item We analyze architectural and governance implications of introducing AI-mediated access into regulated data ecosystems.
\end{enumerate}

\section{Limitations of Existing Integration Approaches}

Although Data Spaces and LLM tool invocation have both received significant attention, their intersection remains unaddressed. data space  platforms target structured protocol-driven interactions \cite{ref5}, whereas LLM frameworks address simpler invocation scenarios \cite{ref2}.

Currently, there is a lack of standardized open-source architectural pattern formalizing LLM mediation over data space  infrastructures, nor empirical evidence demonstrating MCP-based integration \cite{ref14}. Furthermore, governance and security implications of AI-mediated data space  access have not been systematically analyzed. This paper addresses these gaps.

We evaluated direct integration strategies. One approach involves exposing data space  REST APIs directly as tools accessible to LLM agents. However, data space  APIs assume explicit awareness of protocol semantics and multi-step negotiation workflows. Direct exposure risks incorrect invocation patterns and increases coupling between AI agents and low-level protocol details.

Another alternative considered embedding AI-specific extensions directly within the Eunomia Agent. While technically feasible, this strategy would tightly couple a standards-based data space  implementation with rapidly evolving AI tooling ecosystems \cite{ref15}. Such coupling introduces long-term maintainability and compliance risks.

Automated wrapper generation based on API specifications was also examined. Although this approach simplifies interface creation, it does not resolve the underlying semantic gap between declarative tool invocation and protocol-governed interactions.

These limitations motivated the introduction of a dedicated Architectural Mediation Layer to provide controlled abstraction while preserving data space  integrity.

\section{Architectural Mediation Layer}

To reconcile declarative LLM interaction models with procedural data space  workflows, we introduce an Architectural Mediation Layer positioned between MCP clients and the Eunomia Agent. This layer implements MCP tool advertisement and invocation semantics while translating requests into compliant data space  API calls.

The mediation layer operates as an external server communicating with the Eunomia Agent exclusively through public HTTP APIs. It does not embed data space  specific business logic or replicate negotiation workflows. Instead, it abstracts selected capabilities, specifically catalog discovery and data service invocation, into structured MCP tools.

This architectural approach differs from conventional API gateways. While gateways forward requests and manage routing, the mediation layer abstracts protocol semantics into a schema driven interface tailored for LLM consumption. It establishes a stable interoperability boundary that decouples AI evolution cycles from data space  standardization processes.

A central design decision was to ensure that the mediation layer does not replicate data space  governance logic. All identity verification, contract negotiation, and policy enforcement mechanisms remain authoritative within the Eunomia Agent. This separation of concerns reduces duplication, simplifies compliance auditing, and limits the impact of future AI tooling changes on the data space  core. From a software architecture perspective, the mediation layer establishes a stable interoperability boundary that isolates evolving AI interfaces from standards-driven infrastructures.

\section{Architecture Overview}

The proposed architecture consists of three components, as illustrated in Fig.~\ref{fig:combined}: an MCP client, the Mediation Layer server, and the Eunomia Agent. The design follows a non intrusive approach: the data space  implementation remains unchanged and is accessed exclusively through public HTTP APIs.

\subsection{System Components}

The architecture consists of three main components:

\textbf{MCP Client.} The MCP client may be any MCP-compatible LLM environment capable of discovering tools and invoking them through structured schemas. The LLM interacts with external systems solely through MCP, without data space -specific protocol knowledge \cite{ref8}.

\textbf{Mediation Layer Server.} The mediation layer server implements MCP using JSON-RPC over standard input/output transport \cite{ref8}. It exposes a fixed set of MCP tools, abstracting data space  capabilities. Upon initialization, it advertises available tools and their schemas. When a tool is invoked, the server validates input parameters, translates them into HTTP requests targeting the Eunomia Agent, and normalizes responses into MCP-compliant outputs.

\textbf{Eunomia Agent.} A full-fledged data space  agent providing catalog management, dataset metadata, and data service registration. The Eunomia Agent remains unchanged and provides catalog and metadata services through its public REST APIs. Interaction follows a deterministic request–response sequence: tool discovery, dataset listing, metadata retrieval, and data service invocation.

\subsection{Interaction Flow}

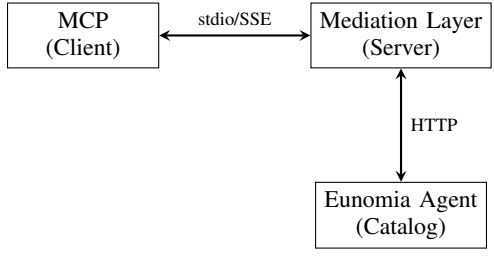
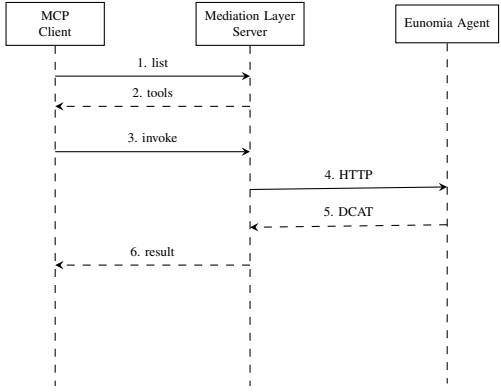
\begin{figure}[!t]
\centering
\begin{subfigure}[t]{0.48\textwidth}
\centering
\begin{tikzpicture}[
    node distance=1.5cm,
    box/.style={rectangle, draw, minimum width=2cm, minimum height=0.8cm, align=center, font=\small},
    arrow/.style={->, >=stealth, thick}
]
    \node[box] (client) {MCP\\(Client)};
    \node[box, right=2cm of client] (server) {Mediation Layer\\(Server)};
    \node[box, below=1.5cm of server] (rainbow) {Eunomia Agent\\(Catalog)};
    
    \draw[arrow, <->] (client) -- node[above, font=\scriptsize] {stdio/SSE} (server);
    \draw[arrow, <->] (server) -- node[right, font=\scriptsize] {HTTP} (rainbow);
\end{tikzpicture}
\caption{Architecture overview}
\label{fig:architecture_a}
\end{subfigure}

\vspace{0.5cm}

\begin{subfigure}[t]{0.48\textwidth}
\centering
\begin{tikzpicture}[
    node distance=0.5cm,
    component/.style={rectangle, draw, minimum width=1.3cm, minimum height=0.5cm, align=center, font=\tiny},
    message/.style={->, >=stealth, font=\tiny},
    return/.style={->, >=stealth, dashed, font=\tiny}
]
    \node[component] (client) {MCP\\Client};
    \node[component, right=1.2cm of client] (server) {Mediation Layer\\Server};
    \node[component, right=1.2cm of server] (rainbow) {Eunomia Agent};
    
    \draw[dashed] (client.south) -- ++(0,-4.5);
    \draw[dashed] (server.south) -- ++(0,-4.5);
    \draw[dashed] (rainbow.south) -- ++(0,-4.5);
    
    \draw[message] ([yshift=-0.4cm]client.south) -- node[above] {1. list} ([yshift=-0.4cm]server.south);
    \draw[return] ([yshift=-0.8cm]server.south) -- node[above] {2. tools} ([yshift=-0.8cm]client.south);
    
    \draw[message] ([yshift=-1.4cm]client.south) -- node[above] {3. invoke} ([yshift=-1.4cm]server.south);
    \draw[message] ([yshift=-1.9cm]server.south) -- node[above] {4. HTTP} ([yshift=-1.9cm]rainbow.south);
    \draw[return] ([yshift=-2.4cm]rainbow.south) -- node[above] {5. DCAT} ([yshift=-2.4cm]server.south);
    \draw[return] ([yshift=-2.9cm]server.south) -- node[above] {6. result} ([yshift=-2.9cm]client.south);
    
\end{tikzpicture}
\caption{Interaction flow sequence showing MCP protocol handshake, tool invocation, and data space  API interaction.}
\label{fig:interaction_b}
\end{subfigure}

\caption{(a) Architecture overview and (b) MCP interaction flow sequence.}
\label{fig:combined}
\end{figure}

The interaction follows a request–response pattern illustrated in Fig.~\ref{fig:interaction_b}.

The interaction proceeds as follows:

\begin{enumerate}
\item The MCP client connects to the mediation layer server and requests available tools through MCP discovery.
\item The mediation layer server advertises supported tools with input/output schemas.
\item Upon tool invocation, the mediation layer server translates requests into HTTP calls to Eunomia agent API endpoints.
\item Eunomia agent processes requests and returns catalog metadata or data service information.
\item The mediation layer server transforms responses into MCP format and relays them to the client.
\end{enumerate}

The prototype implementation of the mediation layer server employs a local stdio transport between the LLM runtime and the mediation server in order to simplify experimentation and deterministic testing. This configuration assumes a trusted execution environment and should be interpreted as a development deployment model rather than a production architecture.

In enterprise scenarios, the mediation server is intended to operate as an independently deployable component, either as a sidecar service co-located with an AI runtime or as a remote mediation service exposed through secure transports such as Server-Sent Events (SSE) or HTTPS-based MCP communication. Under this deployment model, authentication, service identity, and transport-level security mechanisms enforce trust boundaries between the agent environment and data space  infrastructure, enabling distributed cloud-native operation while preserving governance guarantees.

For data access, the server invokes HTTP endpoints associated with selected services, passing query parameters from tool input and relaying returned data without further processing.

This approach prioritizes modularity and separation of concerns. The server acts solely as a translation layer between MCP tool invocations and data space  API calls. The mediation server does not implement data space  business workflows or contract negotiation logic. Instead, it operates as a semantic translation layer that interprets standardized metadata vocabularies, such as DCAT-AP, to dynamically resolve service endpoints and expose them as structured agent tools. This distinction allows the component to remain infrastructure-oriented while avoiding coupling with domain-specific negotiation or governance processes.

This pattern allows organizations to introduce AI agents incrementally without redesigning existing data space  infrastructures, reducing adoption risk while preserving compliance guarantees.

\section{Mediation Layer Implementation}

The Mediation layer implementation emphasizing non-intrusive integration and strict MCP compliance \cite{ref8}. Implemented as a standalone service, it exposes a fixed set of MCP tools corresponding to data space  capabilities without assumptions about the platform's internal architecture beyond public HTTP APIs.

\subsection{Implementation Details}

The server implements the standard MCP interaction model using JSON-RPC over stdio transport \cite{ref8}. This enables seamless integration with MCP-compatible LLM clients without network configuration or service discovery.

Upon initialization, the server advertises supported tools and input schemas. Tool invocation requests are dispatched to dedicated handlers responsible for input validation, data space  API invocation, and MCP-compliant response formatting. Configuration parameters, such as the data space  provider base URL, are externalized via environment variables to support flexible deployment and reproducibility.

\subsection{Exposed MCP Tools}

The prototype exposes three MCP tools enabling dataset discovery and data service invocation, summarized in Table~\ref{tab:tools_combined}.

\textbf{list\_datasets().} Discovers datasets registered in the data space  catalog by querying catalog APIs and aggregating metadata. Responses include essential attributes: dataset identifiers, titles, and descriptions, facilitating LLM-driven selection.

In large-scale Data Spaces, dataset catalogs may contain thousands of entries, exceeding the practical context limits of contemporary LLMs. To address this constraint, the mediation layer exposes catalog discovery through queryable and paginated interactions rather than full catalog retrieval. The list\_datasets tool therefore supports filtering parameters such as keyword queries, semantic categories, and pagination controls (e.g., limit and offset), allowing the agent to iteratively refine discovery requests.

This design prevents context saturation while aligning agent interaction with incremental information retrieval patterns commonly used in enterprise data platforms.

\textbf{get\_dataset\_metadata.} Retrieves detailed metadata for a selected dataset, including associated data service information. Exposes descriptive attributes and endpoint information required for data access operations.

\textbf{query\_data\_service.} Invokes a data service associated with a dataset, mapping structured input parameters to HTTP query parameters and invoking endpoints specified in dataset metadata. Returned data is relayed to the MCP client without further transformation.

Together, these tools provide a minimal yet functional interface for evaluating MCP-based data space  access.

\subsection{API Mapping}

Each MCP tool maps to one or more public HTTP API endpoints exposed by the data space  platform, as detailed in Table~\ref{tab:tools_b}. Mapping logic is implemented explicitly within the server, ensuring transparency and traceability. Catalog tools rely exclusively on catalog APIs, while data access operations invoke HTTP endpoints registered in data service metadata. This approach preserves data space  API semantics while presenting a uniform MCP interface tailored to LLM consumption.

\begin{table}[t]
\centering
\caption{MCP Tools and Their Mapping to Eunomia API Endpoints}
\label{tab:tools_combined}
\tiny

\begin{subtable}[t]{0.48\textwidth}
\centering
\caption{MCP Tools Exposed by the Server}
\label{tab:tools_a}
\begin{tabular}
{p{2.2cm}p{1.3cm}p{1.3cm}p{1.5cm}}
\hline
\textbf{Tool} & \textbf{Input} & \textbf{Output} & \textbf{Purpose} \\
\hline
\texttt{list\_datasets()} & query, limit, offset & Dataset list & Catalog discovery \\
\texttt{get\_dataset\_metadata} & dataset\_id & Metadata & Detail retrieval \\
\texttt{query\_data\_service} & service\_id, params & JSON data & Data access \\
\hline
\end{tabular}
\end{subtable}
\hfill
\begin{subtable}[t]{0.48\textwidth}
\centering
\caption{MCP Tool to Eunomia API Endpoint Mapping}
\label{tab:tools_b}
\begin{tabular}{p{2.3cm}p{0.8cm}p{3.5cm}}
\hline
\textbf{MCP Tool} & \textbf{Method} & \textbf{Eunomia API Endpoint} \\
\hline
\texttt{list\_datasets} & GET & \texttt{/catalogs/\{id\}/datasets} \\
\texttt{get\_dataset\_metadata} & GET & \texttt{/datasets/\{id\}} \\
 & GET & \texttt{/datasets/\{id\}/distributions} \\
\texttt{query\_data\_service} & GET & \texttt{\{service\_endpoint\}?params} \\
\hline
\end{tabular}
\end{subtable}

\end{table}

Implementation source code is publicly available as open-source software: the Eunomia data space  agent \cite{ref12} and the MCP server for Eunomia integration \cite{ref13}.

\section{Proof of Concept and Evaluation}

\subsection{Evaluation Setup}

The evaluation environment consists of a Eunomia Agent provider instance, the MCP mediation server, a deterministic MCP client, and a mock HTTP data service. The mock service simulates data retrieval without introducing backend complexity. A dedicated dataset titled “MCP Evaluation Dataset” is registered in the Eunomia Agent catalog and linked to the mock service via DCAT-AP 3.0 metadata.

\textbf{Eunomia data space  Provider.} A full Eunomia agent instance exposing its catalog through HTTP APIs, managing datasets and data services following DCAT-AP 3.0. Although Eunomia supports additional capabilities, including Self-Sovereign Identity and dataplane coordination, only catalog publication and metadata discovery are exercised, consistent with the proof-of-concept scope.

\textbf{MCP Mediation Server.} Implemented as an external standalone service communicating with Eunomia exclusively through public REST APIs. The server exposes three MCP tools (list\_datasets, get\_dataset\_metadata, query\_data\_service) using JSON-RPC over stdio transport.

\textbf{MCP Evaluation Client.} A deterministic client executing a predefined sequence of MCP calls without LLM integration or semantic reasoning. This design ensures reproducibility and isolates evaluation from language model variability.

\textbf{Mock Data Service.} A lightweight HTTP service registered in Eunomia's catalog, accepting GET requests with query parameters and returning synthetic JSON records. This completes the end-to-end flow without introducing real data backend complexity.

A dedicated catalog entry is created programmatically in Eunomia: a dataset titled "MCP Evaluation Dataset" with an associated data service and DCAT-AP 3.0 distribution linking them via dcat:accessService. All components communicate locally, enabling deterministic single-machine reproduction.

\subsection{Evaluation Methodology}

The evaluation follows a deterministic four-step flow: First, the client requests tool discovery to verify MCP compliance and server initialization. Second, it invokes list\_datasets and deterministically selects the first returned dataset. Third, it invokes get\_dataset\_metadata and selects the first associated data service. Fourth, it invokes query\_data\_service with fixed query parameters. This sequence validates functional correctness, protocol compliance, and integration feasibility rather than performance or semantic quality.

\textbf{Tool Discovery.} The client invokes MCP tools/list to retrieve the advertised tools, validating server reachability, correct tool advertisement, and a successful protocol handshake.

\textbf{Dataset Listing.} The client invokes list\_datasets(), which the server translates into Eunomia catalog API calls. The client deterministically selects the first returned dataset without filtering or semantic criteria.

\textbf{Dataset Metadata Retrieval.} The client invokes get\_dataset\_metadata using the selected dataset identifier. The server retrieves detailed metadata, including associated data services via DCAT-AP 3.0 distributions. The client deterministically selects the first available data service.

\textbf{Data Service Query.} The client invokes query\_data\_service with the selected service identifier and fixed query parameters. The server resolves the endpoint URL from Eunomia metadata and issues an HTTP request. The response is returned to the client as JSON.

This fixed sequence ensures full reproducibility. The evaluation success requires that each step completes without errors, that the metadata returned corresponds to registered catalog entities, and that the data service invocation returns structured JSON data. Worded differently, it succeeds if: (1) all MCP protocol interactions complete without errors; (2) each tool invocation produces semantically consistent responses; (3) retrieved metadata corresponds to Eunomia catalog entities; (4) the data service endpoint is successfully invoked with a JSON response returned. Failure at any step constitutes evaluation failure.

The methodology deliberately excludes semantic interpretation and dynamic tool selection, focusing on validating the MCP abstraction layer for data space  discovery and invocation.

\subsection{Evaluation Results}

To assess practical viability, an integration scenario was implemented covering catalog discovery, metadata retrieval, and data service invocation. The objective was to validate architectural coherence and interaction correctness across system boundaries rather than to benchmark performance.

The results show that data space  capabilities can be exposed as structured MCP tools without modifying the Eunomia Agent codebase. The mediation layer successfully translated catalog queries into normalized dataset metadata, resolved dataset-to-service relationships using Eunomia information models, and executed data service invocations through dynamically discovered endpoints. All interactions completed without protocol inconsistencies or integration errors, confirming reliable end-to-end operation across the mediation workflow.

The complete interaction chain from MCP client through the mediation layer and Eunomia services to the data endpoint operated as intended, demonstrating interoperability between agent-based environments and data space  infrastructures. Integration with Claude Desktop further validated compatibility with production LLM environments, where the client discovered available tools, interpreted their schemas, and invoked them through natural language interaction without additional configuration.

Rather than demonstrating isolated functionality, these results indicate that protocol-based mediation enables controlled exposure of data space  capabilities to LLM agents while preserving governance boundaries. This validation establishes the architectural feasibility of introducing AI agents as managed participants within data space  ecosystems, providing the foundation for the broader implications discussed in the following section.

\section{Security and Governance Considerations}

The mediation layer operates at the intersection of AI autonomy and regulated data space  infrastructures, serving as a strategic point for enforcing governance, organizational policies, and regulatory requirements. Safeguards such as explicit allowlists for tool invocation, strict schema validation, and parameter sanitization are essential to ensure that AI driven automation respects contractual data usage constraints \cite{ref18}. By centralizing access to data space  capabilities, the layer establishes clear trust boundaries between LLM clients and the data space  platform.

In the current proof of concept implementation, the server operates locally via stdio transport without authentication, relying on process isolation for security. It functions as a controllable enforcement point where validation, auditing, and authorization checks can be integrated. By intercepting tool invocations before they reach the data space  platform, the mediation layer enables centralized policy enforcement without requiring changes to LLM clients or data space  backends. However, exposing data space  capabilities as MCP tools introduces risks. Overly permissive tool definitions or lack of authentication could allow unintended access, while LLM agents remain vulnerable to prompt injection attacks that manipulate behavior or exfiltrate data. Catalog metadata itself may reveal sensitive information about organizational structure, even when direct data access is restricted.

Future integration Self Sovereign Identity (SSI) offers a path to robust authentication and per invocation authorization. By verifying SSI credentials at the mediation layer, tool level access can be restricted based on identity, usage policies, and contract terms using secure and optimized protocols \cite{ref16}. This approach mitigates prompt injection risks, enforces metadata-level access control, and ensures compliance with agreed data usage policies. Combined with audit trails capturing credential verification, tool invocations, and authorization decisions, SSI integration would strengthen security and compliance while preserving the mediation layer’s non intrusive design.

\section{Lessons Learned}

Integrating LLM agents with Data Spaces presents architectural challenges that go beyond implementation. A central issue is the asymmetry between rapidly evolving AI tooling ecosystems and the stability focused, standards based data space  infrastructures. Mediation boundaries are essential to shield regulated systems from AI volatility while preserving governance and compliance guarantees.

Tool schemas play a critical role, acting as long term interface contracts where even minor changes can affect reasoning and invocation patterns. Effective schema governance through version control, validation, and careful change management is, therefore, a core architectural responsibility.

The mediation layer itself serves not merely as a translation component but as a governance control point. By centralizing access to data space  capabilities, it enables authentication, rate limiting, logging, auditing, and parameter validation, strengthening both resilience and compliance.

AI-driven access also introduces unique security challenges. Autonomous tool selection and prompt driven invocations create new threat vectors, requiring defensive interface design, strict parameter sanitization, and clearly defined trust boundaries. Successfully integrating AI agents into Data Spaces requires a governance approach that addresses both the technical and behavioral dimensions of system interaction. Beyond validating technical feasibility, the integration process revealed broader architectural insights relevant to enterprise practitioners.

\section{Conclusion and Future Work}

The integration of large language model agents into Data Spaces introduces architectural challenges that extend beyond technical interoperability, requiring mechanisms capable of reconciling probabilistic AI behavior with policy-driven data exchange infrastructures. This article presented an architectural mediation approach based on the Model Context Protocol and implemented through the Eunomia Agent, enabling controlled interaction between LLM-based agents and data space  services while preserving governance guarantees.

The implemented prototype demonstrated the feasibility of this mediation layer through end-to-end interaction scenarios, confirming that standardized protocol translation can reliably expose data space  capabilities as structured tools consumable by LLM agents. Rather than representing an isolated technical integration, these results illustrate how architectural mediation can transform AI agents from experimental interfaces into governed participants within enterprise data ecosystems.

From an organizational perspective, the proposed approach reduces adoption barriers for AI-driven automation by allowing enterprises to introduce intelligent agents incrementally without redesigning existing data space  infrastructures. This separation between AI interaction logic and data space  governance mechanisms enables organizations to maintain compliance, traceability, and control while benefiting from emerging agent-based workflows. As Data Spaces continue to evolve as foundational components of digital sovereignty initiatives, such mediation architectures may become essential integration patterns for enterprise AI deployment.

Future work will focus on extending the mediation model toward multi-agent coordination, dynamic policy negotiation, and adaptive governance mechanisms capable of responding to evolving regulatory and operational requirements. Further evaluation in large-scale industrial deployments will also be necessary to assess long-term scalability and operational sustainability.

\section*{Acknowledgments}
This work was partially supported by the Spanish Agencia Estatal de Investigación under Grants FUN4DATE (PID2022-136684OB-C22) and SMARTY (PCI2024-153434), by TUCAN6-CM (TEC-2024/COM460) funded by CM (ORDEN 5696/2024) and SMARTY funded by the European Commission through the Chips Act Joint Undertaking project SMARTY (Grant 101140087).

\end{document}